\documentclass[conference]{IEEEtran}

\usepackage{algorithm}
\usepackage{algorithmic}

\makeatletter
\newcommand\fs@norules{
\def\@fs@cfont{\bfseries}\let\@fs@capt\floatc@ruled  \def\@fs@pre{}  \def\@fs@post{}  \def\@fs@mid{\kern3pt}  \let\@fs@iftopcapt\iftrue  }
\makeatother
\floatstyle{norules}
\restylefloat{algorithm}

\newlength\myindent
\usepackage{blindtext, graphicx}
\usepackage{tikz}
\usetikzlibrary{arrows,shapes, calc,decorations.text,decorations.pathreplacing}

\usepackage{cite}

\ifCLASSINFOpdf
\else
\fi
\usepackage{amssymb}
\usepackage[cmex10]{amsmath}
\usepackage{algorithmic}
\newcommand{\ocap}{\operatornamewithlimits{\text{\textcircled{\scalebox{1.4}{\tiny{$\cap$}}}}}}

\usepackage{array}
\usepackage{color}
\begin{document}
%
\title{Preference fusion and Condorcet's Paradox under uncertainty}

\author{\IEEEauthorblockN{Yiru Zhang}
\IEEEauthorblockA{IRISA - University of Rennes 1\\
France\\
Email: yiru.zhang@irisa.fr}
\and
\IEEEauthorblockN{Tassadit Bouadi}
\IEEEauthorblockA{IRISA - University of Rennes 1\\
France\\
Email: tassadit.bouadi@irisa.fr}
\and
\IEEEauthorblockN{Arnaud Martin}
\IEEEauthorblockA{IRISA - University of Rennes 1\\
France\\
Email: arnaud.martin@irisa.fr}}


%


\maketitle

\begin{abstract}
Facing an unknown situation, a person may not be able to firmly elicit
his/her preferences over different alternatives, so he/she tends to express uncertain preferences. Given a community of different persons expressing their preferences over certain alternatives under uncertainty, to get a collective representative opinion of the whole community, a preference fusion process is required.
The aim of this work is to propose a preference fusion method that copes with uncertainty and escape from the Condorcet paradox. To model preferences under uncertainty, we propose to develop a model of preferences based on belief function theory that accurately describes and captures the uncertainty associated with individual or collective preferences. This work improves and extends the previous results.
This work improves and extends the contribution presented in a previous work. 
The benefits of our contribution are twofold. On the one hand, we propose a qualitative and expressive preference modeling strategy based on belief-function theory which scales better with the number of sources. On the other hand, we propose an incremental distance-based algorithm (using Jousselme distance) for the construction of the collective preference order to avoid the Condorcet Paradox.
\end{abstract}


%
\IEEEpeerreviewmaketitle

\section{Introduction}\label{sec:intro}

The concept of preferences has been widely studied in the domain of databases, information systems and artificial intelligence \cite{Lacroix1987}. Expressing preferences over a set of alternatives or items such as music clips or literature books may reveal the education level, personal interests or other interesting user's informations. Various applications based on preferences have emerged in recommendation systems~\cite{Xiang2010}, rank manipulation in statistics~\cite{Marden1995}, decision making, community detection, etc. However, preferences are not always expressed firmly, sometimes a preference may be uncertain facing an unknown situation so that the relation between two items can be ambiguous.
Tony is asked to express his preference between ``\textit{apple}" and ``\textit{ramboutan}". However, Tony does know how \textit{ramboutan} tastes. By referring to the picture of \textit{ramboutan}, Tony finds that it's similar to \textit{litchi} in terms of the form and Tony usually prefers \textit{litchi} to \textit{apple}. With such knowledge, Tony decided to state that he prefers \textit{ramboutan} to \textit{apple} for 80\% of certainty and prefers \textit{apple} for the rest 20\%. Kevin has the same knowledge level as Tony and is asked to compare \textit{litchi} to \textit{ramboutan}. So Kevin gives a high value 70\% on ``indifference" relation.
 Given a community of different users expressing their preferences over certain items under uncertainty, to get a collective representative opinion of the whole community, a preference fusion process is required.

Various paradoxes and impossibility theorems may arise from a preference fusion process, and one of the most well-known is Condorcet Paradox~\cite{Condorcet1785}. Condorcet paradox, also known as voting paradox, is a situation in which collective preferences can be cyclic ({\em i.e.} not transitive) even the preferences of individual voters are transitive. 
Let's imagine a situation where Tony prefers \textit{apple} to \textit{pear}, Kevin prefers \textit{pear} to \textit{orange} and David prefers \textit{orange} to \textit{apple}. After the fusion of their individual transitive preferences, 3 items \textit{apple, pear, orange} form the following relationship: \textit{apple} is preferred to \textit{pear}, \textit{pear} is preferred to \textit{orange} and \textit{orange} is preferred to \textit{apple}. In this situation, we can notice that there is a violation of transitivity in the collective preference ordering. 
The aim of this work is to propose a preference fusion method that copes with uncertainty and escapes from the Condorcet paradox. This work improves and extends the contribution presented in \cite{elarbi:hal-01270169}.
The benefits of our contribution are twofold. On the one hand, we propose a qualitative and expressive preference modeling strategy based on belief-function theory. On the other hand, we propose an incremental construction of the collective preference order that avoid the Condorcet Paradox. 

In this paper, we use the term ``agent" to refer ``person", ``user" or ``voter", and term ``alternative" for ``item", ``candidate" or ``object".

The rest of the paper is organized as follows.
In section \ref{sec:stateoftheart}, we discuss related work on: (i) the problem of preference aggregation under uncertainty, and (ii) the Condorcet's Paradox phenomenon. Section~\ref{sec:concepts} recalls basic concepts related to preference orders and belief functions. The details of our method is explained in Section \ref{sec:preferencefusion}. Experiments and their analysis are given in Section \ref{sec:experiments} followed by some conclusive discuss in Section \ref{sec:conclusion}. 

\section{State-of-the-art}\label{sec:stateoftheart}
The representation of preferences has been studied in various domains such as decision theory~\cite{Ozturke2005}, artificial intelligence~\cite{Wellman91preferentialsemantics}, economics and sociology~\cite{arrow1959rational}. Preferences are essential to efficiently express user's needs or wishes in decision support systems such as recommendation systems and other preference-aware interactive systems that need to elicit and satisfy user preferences. However, preference modeling and preference elicitation are not easy tasks, because human beings tend to express their opinions in natural language rather than in the form of preference relations. 
Preferences are also widely used in collective decision making and social choice theory, where the group's choice is made by aggregating individual preferences.

In this paper, we address two major issues. The first one concerns the problem of aggregation of mono-criterion preference relations under uncertainty. In fact, the aggregation of mono-criterion preferences for collective decision making is a very traditional problem, also known as voting problem. Studies on different electoral voting systems can be found in \cite{cox1997making}. The majority of voting methods require voters to make binary comparisons, declaring that one alternative is preferred to another one, without taking into account uncertainty in the preference relationship. 
Belief function theory is a mathematical framework for representing and modeling uncertainty \cite{COIN:COIN244}. In \cite{Denoeux2012, schubert2014partial}, the authors proposed a belief-function-based model for preference fusion, allowing the expression of uncertainty over the lattice order ({\em i.e.} preference structure). However, this modeling approach does not constitute an optimal representation of preferences in the presence of uncertain and voluminous information.
Based on \cite{Denoeux2012}, the model of uncertainty proposed by Masson, {\em et al.} in \cite{Masson2016} allows the expression of uncertainty on binary relations ({\em i.e.} preference relations). Precisely, this approach proposes to model preference uncertainty over binary relations between pairs of alternatives, and to infer partial orders. The preference relations considered in this method are : ``strict preference" and ``incomparability". The``indifference" relation has not been addressed. Furthermore, this approach defines only one mass function on each alternative pair, which is not expressive enough. In fact, this does not allow to express uncertainty over ``strict preference" and ``incomparability" at the same time.

The work proposed in \cite{elarbi:hal-01270169} improved the method introduced in \cite{Masson2016} by considering the ``indifference" relation. The authors proposed to define belief degrees on each alternative pair to be compared, and to interpret this degrees as elementary belief masses. Nonetheless, the combination method in \cite{elarbi:hal-01270169} does not scale with the number of information sources (agents).

The second issue we should tackle is the problem of Condorcet Paradox ({\em i.e.} Condorcet Cycles). The Condorcet Paradox states that aggregating transitive individual preferences can lead to intransitive collective preferences. To overcome this problem, a known method is to consider the size of majorities supporting various collective preferences, by breaking the link of the cycle with the lowest number of supporters~\cite{Nurmi2001}. This method works only on non-valued structures and cannot be applied to our case ({\em i.e.} preference fusion under uncertainty).
Furthermore, most of the work considering preference fusion under uncertainty \cite{Masson2016, Denoeux2012,elarbi:hal-01270169} does not address the problem of Condorcet Paradox. 

We therefore propose, as an extension work of Elarbi, {\em et al.} \cite{elarbi:hal-01270169}, a preference fusion method based on belief function theory~\cite{shafer1976mathematical}, allowing a numeric expression of uncertainty. We also propose a preference graph construction method combining Tarjans's strongly connected components algorithm~\cite{Tarjan72depthfirst} and Jousselme distance~\cite{Jousselme200191} to avoid Condorcet paradox~\cite{Condorcet1785}. 
\section{Notation and Basic Definitions}
\label{sec:concepts}
 In this section, we present the necessary concepts and definitions related to preference orders and in belief function theory. Many are borrowed from ~\cite{Ozturke2005}.

\subsection{Preference Order}
\newtheorem{mydef}{DEFINITION}
\begin{mydef}\label{def:binaryrelation}
(Binary Relation) Let $A$ be a finite set of alternatives ($a,b,c,\ldots,n$), a \textbf{binary relation} $R$ on the set $A$ is a subset of the cartesian product $A\times A$, that is, a set of ordered pairs ($a,b$) such that $a$ and $b$ are in $A:R\subseteq A\times A$~\cite{Ozturke2005}.
\end{mydef}
A binary relation on set $A$ can be represented by a graph or an adjacency matrix~\cite{Ozturke2005}. A binary relation can satisfy any of the following properties: \textit{reflexive, irrelfexife, symmetric, antisymmetric, asymmetric, complete, stronly complete, transitive, negatively transitive, semitransitive, and Ferrers relation}~\cite{Ozturke2005}. The detailed definitions of the properties are not in the scope of this article. 

Based on definition \ref{def:binaryrelation}, we denote the binary relation ``prefer" by $\succeq$. $x\succeq y$ means ``$x$ is at least as good as $y$". 
Inspired by a four-valued logic introduced in~\cite{Belnap1977,Masson2016,elarbi:hal-01270169}, we introduce four relations between alternatives. Given a set $A=\{a, b,c,\ldots,n\}$ and a preference order $\succeq$ defined on $A$, we have $\forall a,b \in A$, three possible relations, strict preference, indifference and incomparability, may exist. They are defined respectively by:
\begin{itemize}
\item \textbf{Strict preference} denoted by $P$: $a\succ b$\\ ($a$ is strictly preferred to $b$)\\
$\Leftrightarrow a\succeq b\wedge \neg (b\succeq a)$
\item \textbf{Inverse strict preference} denoted by $\neg P$: $b\succ a$\\ ($a$ is inversely strictly preferred to $b$)\\
$\Leftrightarrow \neg (b\succeq a)\wedge b\succeq a$
\item \textbf{Indifference} denoted by $I$: $a\approx b$\\ ($a$ is indifferent, or equally preferred, to $b$) \\
$\Leftrightarrow a\succeq b\wedge b\succeq a$
\item \textbf{incomparability} denoted by $J$: $a\sim b$\\ ($a$ is incomparable to $b$)$\Leftrightarrow \neg (a\succeq b)\wedge \neg (b\succeq a)$
\end{itemize}

A preferences structure $\langle P,I,J \rangle$ on multiple alternatives can therefore be presented by a binary relation~\cite{Ozturke2005}. 
\begin{mydef}
\label{def:preferencestructure}
(Preference Structure) A preference structure is a collection of binary relations defined on the set $A$ such that for each pair $a, b$ in $A$:
\begin{itemize}
\item at least one relation is satisfied
\item if one relation is satisfied, another one cannot be satisfied.
\end{itemize}
\end{mydef}

The preference order is a partial order~\cite{Ozturke2005}, defined as:
\begin{mydef}
\label{def:partialorder}
(Quasi Order) Let $R$ be a binary relation ($R=P \cup I$) on the set $A$, $R$ being a characteristic relation of $\langle P , I , J \rangle$, the following three definitions are equivalent:
\begin{enumerate}
\item $R$ is a quasi order.
\item $R$ is reflexive, transitive.
\item  \[\left\{
                \begin{array}{ll}
                  P$ is asymmetric, transitive$\\
                  I$ is reflexive, symmetric$\\
                  J$ is irreflexive and symmetric$\\
                  (P.I\cup I.P)\subset P.
                \end{array}
              \right.
              \]
              
\end{enumerate}
\end{mydef}
\subsection{Belief functions}
The theory of belief functions (also referred to as Dempster-Shafer or Evidence Theory) was firstly introduced by Dempster~\cite{dempster1967} then developped by Shafer \cite{shafer1976mathematical} as a general model of uncertainties. It is applied widely in information fusion and decision making. By mixing probabilistic and set-valued representations, it allows to represent degrees of belief and incomplete information in a unified framework.
Let $\Omega=\{\omega_1,\ldots\omega_n\}$ be a finite set. A \textit{(normalized) mass function} on $\Omega$ is a function $m:2^{\Omega} \rightarrow [0,1]$ such that:
\begin{equation}
m(\emptyset)=0
\end{equation} 
\begin{equation}
\sum_{\substack{X\subseteq\Omega}}m(X)=1
\end{equation}

The subsets $X$ of $\Omega$ such that $m(X)>0$ are called \textit{focal elements} of \textit{m}, while the finite set $\Omega$ is called \textit{framework of discernment}. 
A mass function is called \textit{simple support} if it has only two focal elements: $X\subseteq \Omega$ and $\Omega$.
In this paper, we consider only the normalized mass functions.\\
With the definition of mass function, information from different sources can be fused based on different combination rules. Two types of combination rules are often applied on cognitively independent source (definition~\ref{def:cognitive}): conjunctive combination~\cite{Smets1990} and disjunctive combination. Here we only introduce the former one which is applied in our work.
\begin{mydef}
\label{def:cognitive}
(Cognitively Independence Source) sources are considered as cognitively independent if the belief on any one source has no communication with the others.
\end{mydef}
In case that the mass functions are not cognitively independent, we apply a combination rule taking the mean value defined by Denoeux~\cite{Denoeux2006}. For \textit{n} non-normalized and non-independent mass functions, the mean result is:
\begin{equation}\label{eq:meancombine}
m_{mean} (X) = \dfrac{1}{S}\displaystyle\sum_{s=1}^{S}m_s(X)
\end{equation}
 
The conjunctive combination rule proposed by~\cite{Smets1990} is applied for finding the consensus among multiple reliable and cognitively independent sources. Given a discernment framework $\Omega$ and sources $S$, mass function on source $s \in S$ denoted as $m_s$, for $\forall X \in 2^\Omega$, we have:
\begin{equation}\label{eq:conjcombine}
m_{conj}(X) = (\displaystyle \ocap_{s=1}^{S} m_s)(X) = \displaystyle\sum_{Y_1\cap \ldots \cap Y_s = X}\prod_{s=1}^{S} m_s(Y_s)
\end{equation}

In information fusion process, the last step is decision making, which is to take an element $\omega_i$ or a disjunctive set from $\Omega$ upon the final mass function. Notions of Belief Function $bel$ and plausibility function $pl$ represented by mass functions are often used in this process. $bel$ and $pl$ are given by:
\begin{equation}
bel(X)\triangleq \displaystyle\sum_{Y\subseteq X}m(Y)
\end{equation}
\begin{equation}
pl(X)\triangleq \displaystyle\sum_{Y\cap X \neq \emptyset}m(Y)
\end{equation}

The belief function quantifies how much event $X$ is implied by the information, as it sums up masses of sets included in $X$ and whose mass is necessarily allocated to $X$. The plausibility function quantifies how much event $X$ is consistent with all information, as it sums masses that do not contradict $X$~\cite{Masson2016}. Meanwhile, pignistic probability, a compromised method proposed by Smets~\cite{Smets1990pignistic} is often applied for selection on singletons. For all $X \subseteq \Omega$ pignistic probability $betP$ of $X$ is given by: 
\begin{equation}
betP(X) = \displaystyle\sum_{Y\in 2^{\Omega},Y\cap X\neq \emptyset} \dfrac{1}{\lvert Y \rvert} \dfrac{m(Y)}{1-m(\emptyset)}
\end{equation}
\section{Preference Fusion} \label{sec:preferencefusion}

In this section, we explain our strategies on preference fusion under uncertainty as well as our methods for Condorcet paradox avoidance. To simplify the notations, in this section we represent alternative pairs by their index $(a_i, a_j)$ with the condition that $i<j$.

\subsection{Fusion of mass functions} \label{subsec:fusionstrategies}
According to the theory of belief functions, a typical fusion process includes 3 major steps: \textit{Modeling, Combination and Decision making}. Our approaches are presented in such order.
\subsubsection{Modeling}
We consider the case that a group of $S$ agents expressing their preferences between every pair of alternatives from the set $A = \{a_1,a_2,\ldots,a_n\}$. The framework of discernment is defined on possible relations:
$$\Omega_{ij} = \{\omega_{ij}^1,\omega_{ij}^2,\omega_{ij}^3,\omega_{ij}^4\},$$ 
where $\omega_{ij}^1$, $\omega_{ij}^2$, $\omega_{ij}^3$ and $\omega_{ij}^4$, represent respectively $a_i\succ a_j$, $a_i\prec a_j$, $a_i\approx a_j$ and $a_i\sim a_j$. The objective is to aggregate the preferences of all agents $S$. Each agent $s\in S$ is asked to give a belief degree between 0 and 1 on all of the four relations for each alternative pair  $(a_i,a_j)$. For example, between alternatives $apple$ and $pear$,  agent $s$ gives 4 degrees respectively on $apple \succ pear$, $apple\prec pear$, $apple\approx pear$ and $apple\sim pear$. Hence, the belief degree of agent $s$ is represented by 4 mass functions noted as $m_s^k$, $(k=1,2,3,4)$.

In this model, each mass function in $m_s^{1-4}$ is a simple support and the mass functions are cognitively independent. Such design is more intuitive for the user. Indeed, an individual usually feels more comfortable to give belief degrees on one single event rather than 4 events occurring simultaneously.
\subsubsection{Combination}
In this step, we propose two strategies of combination, namely A and B, for alternative pair indexed by $i,j(i<j)$. The strategies are illustrated in figure~\ref{fig:flowchartA} and \ref{fig:flowchartB} followed by detailed explanations. Strategy A was originally proposed by \cite{elarbi:hal-01270169}. Both strategies are based on the two combination rules in equations~\eqref{eq:meancombine} and \eqref{eq:conjcombine}.

\tikzstyle{decision} = [diamond, draw, fill=blue!20, 
    text width=4.5em, text badly centered, node distance=3cm, inner sep=0pt]
\tikzstyle{block} = [rectangle, draw, 
    text width=5em, minimum height=2em]
\tikzstyle{line} = [draw, -latex']
\begin{itemize}
\item \textbf{Strategy A:}\label{strategyA}
\begin{figure}[h]
\raggedright
\centering
\begin{tikzpicture}
\node [align=left] (sources) at (0,0) {$m_s^1$\\$\ldots$\\$m_s^4$};
\draw [decorate,decoration={brace,amplitude=3pt,mirror,raise=4pt},yshift=0pt]
(0.2,1.1) -- (0.2,2.1) node [black,midway,xshift=0.1cm] (b1) {};
\node [block] (means) at (1.7,0) {mean rule combination};
\node [] (combineds) at (3.5,0) {$m_s^{\Omega_{ij}}$};
\node [align=left] (source1) at (0,1.6) {$m_1^1$\\$\ldots$\\$m_1^4$};
\draw [decorate,decoration={brace,amplitude=3pt,mirror,raise=4pt},yshift=0pt]
(0.2,-0.5) -- (0.2,0.5) node [black,midway,xshift=0.1cm] (bs) {};
\node [block] (mean1) at (1.7,1.6) {mean rule combination (eq~\ref{eq:st1meancomb})};
\node [] (combined1) at (3.5,1.6) {$m_1^{\Omega_{ij}}$};
\node at ($(means)!.5!(mean1)$) {\vdots};
\node at ($(combineds)!.5!(combined1)$) {\vdots};
\node at ($(sources)!.5!(source1)$) {\vdots};
\draw [decorate,decoration={brace,amplitude=3pt,mirror,raise=4pt},yshift=0pt]
(3.75,0) -- (3.75,1.6) node [black,midway,xshift=0.1cm] (bN){};
\node [block] (conjunctive) at (5.3, 0.8) {conjunctive combination (eq~\ref{eq:st1conjcomb})};
\node [] (final) at (7.2,0.8) {$m^{\Omega_{ij}}$};
\path [line] (bs) -> (means);
\path [line] (b1) -> (mean1);
\path [line] (means) -> (combineds);
\path [line] (mean1) -> (combined1);
\path [line] (bN) -> (conjunctive);
\path [line] (conjunctive) -> (final);
\end{tikzpicture}
\caption{Combination Strategy A}
\label{fig:flowchartA} 
\end{figure}
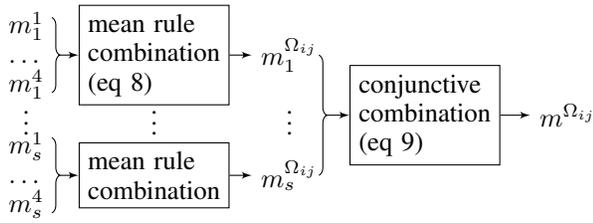

Firstly we combine the four mass functions on four relations of one pair $(a_i,a_j)$ of one agent $s$ into one mass function. Using their mean value upon equation~\eqref{eq:meancombine} as:
\begin{equation}\label{eq:st1meancomb}
m_s^{\Omega_{ij}}(X)=\dfrac{1}{4}\displaystyle\sum_{k=1}^4m_s^k(X)
\end{equation}
Then the conjunctive combination rule in equation~\eqref{eq:conjcombine} is applied over multiple agents. For an alternative pair $(a_i,a_j)$, we obtain a mass function given by:
\begin{equation}\label{eq:st1conjcomb}
m^{\Omega_{ij}}(X) = \displaystyle \ocap_{s=1}^S m_s^{\Omega_{ij}}(X)
\end{equation}

The mass function $m^{\Omega_{ij}}(X)$ is the finally combined mass function for the alternative pair $(a_i,a_j)$. 
One of the drawbacks of strategy A exists in its inability to scale while the volume of information sources (agents in our case) increases. After the first combination with mean value rule, the combined mass functions have non-zero values on the four singletons and are no longer simple support mass functions. The non-simple-support mass functions have auto-conflicts and lead to increase the value of $m^{\Omega_{ij}}(\emptyset)$ when combining its. Once the volume of sources gets large, auto-conflicts are exacerbated by the conjunctive combination rule so that $m^{\Omega_{ij}}(\emptyset)$ converges to 1. To avoid such deterioration, we proposed strategy B in which the conjunctive rule is applied on simple support mass functions.
\item \textbf{Strategy B:}\label{strategyB}
\begin{figure}[h]
\raggedright
\centering
\begin{tikzpicture}
\node [align=left] (source4) at (0,0) {$m_1^4$\\$\ldots$\\$m_s^4$};
\draw [decorate,decoration={brace,amplitude=3pt,mirror,raise=4pt},yshift=0pt]
(0.2,1.1) -- (0.2,2.1) node [black,midway,xshift=0.1cm] (b1) {};
\node [block] (conjunctive4) at (1.7,0) {conjunctive combination};
\node [] (combined4) at (3.3,0) {$m^4$};
\node [align=left] (sources) at (0,1.6) {$m_1^1$\\$\ldots$\\$m_s^1$};
\draw [decorate,decoration={brace,amplitude=3pt,mirror,raise=4pt},yshift=0pt]
(0.2,-0.5) -- (0.2,0.5) node [black,midway,xshift=0.1cm] (b4) {};
\node [block] (conjunctive1) at (1.7,1.6) {conjunctive combination (eq~\ref{eq:st2conjcomb})};
\node [] (combined1) at (3.3,1.6) {$m^1$};
\node at ($(conjunctive1)!.5!(conjunctive4)$) {\vdots};
\node at ($(combined1)!.4!(combined4)$) {\vdots};
\node at ($(sources)!.4!(source4)$) {\vdots};
\draw [decorate,decoration={brace,amplitude=3pt,mirror,raise=4pt},yshift=0pt]
(3.5,0) -- (3.5,1.6) node [black,midway,xshift=0.1cm] (bN){};
\node [block] (mean) at (5.1, 0.8) {mean rule combination (eq~\ref{eq:st2meancomb})};
\node [] (final) at (7,0.8) {$m^{\Omega_{ij}}$};
\path [line] (b4) -> (conjunctive4);
\path [line] (b1) -> (conjunctive1);
\path [line] (conjunctive4) -> (combined4);
\path [line] (conjunctive1) -> (combined1);
\path [line] (bN) -> (mean);
\path [line] (mean) -> (final);
\end{tikzpicture}
\caption{Combination Strategy B}
\label{fig:flowchartB} 
\end{figure}
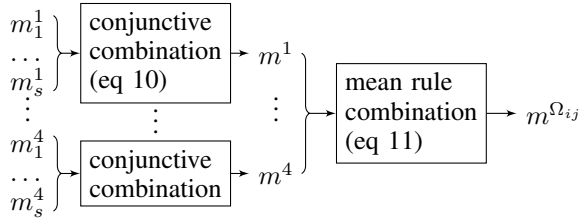

Strategy B is the inverse of strategy A.\\
	Firstly, we cluster mass functions for each alternative pair $(a_i,a_j)$ of all agents $S$ into 4 clusters. On each cluster, the conjunctive combination (equation~\eqref{eq:conjcombine}) is applied.
\begin{equation}\label{eq:st2conjcomb}
	m^{k}(X) = \displaystyle \ocap_{s=1}^S m_s^{k}(X), k=1,\ldots,4
\end{equation}
Then we apply mean value combination method on the 4 combined clusters.
\begin{equation}\label{eq:st2meancomb}
	m^{\Omega_{ij}}(X) = \dfrac{1}{4}\displaystyle \sum_{k=1}^4 m^{k}(X)
\end{equation}
\end{itemize}

\subsubsection{Decision}
After two combinations, we finally get a mass function for each pair $(a_i,a_j)$ denoted by $m^{\Omega_{ij}}$. The decision related to the relationship of each pair is taken based on the pignistic probability on the space $\Omega_{ij}$ by:
\begin{equation}\label{eq:decision}
\omega_{ij}^d = \underset{{\omega_{ij}^k,k={1,\ldots,4}}}{\mathrm{argmax}} betP^{\Omega_{ij}}(\omega_{ij}^k)
\end{equation}
Where $\omega_{ij}^k$ is represented by mass function $m^{\Omega_{ij}}$.
Since the decision is make on aggregated preferences. Conflict preferences such as Condorcet Paradox may appear. In the subsection \ref{subsec:methodcondorcet}, we propose different algorithms to avoid Condorcet Paradox in the final result.

\subsection{Condorcet Paradox Avoidance in Graph Construction}
\label{subsec:methodcondorcet}

In the following, we use directed graph for the graphic representation of the preference order obtained from the fusion process. The four possible relationships are illustrated as follows:
\begin{itemize}
\item $\omega_{ij}^1$: $a_i\succ a_j$: 
\begin{tikzpicture}
\pgfmathsetmacro{\shift}{0.2ex}
\node[draw,circle] (i) at  (0,0) {i};
\node[draw,circle] (j) at  (2,0) {j};
\draw [->] (i) -- (j);
\end{tikzpicture}
\item  $\omega_{ij}^2$: $a_i\prec a_j$:
\begin{tikzpicture}
\pgfmathsetmacro{\shift}{0.2ex}
\node[draw,circle] (i) at  (0,0) {i};
\node[draw,circle] (j) at  (2,0) {j};
\draw [->] (j) -- (i);
\end{tikzpicture}
\item  $\omega_{ij}^3$: $a_i\approx a_j$:
\begin{tikzpicture}
\pgfmathsetmacro{\shift}{0.2ex}
\node[draw,circle] (i) at  (0,0) {i};
\node[draw,circle] (j) at  (2,0) {j};
\draw [->] (i) -- (j);
\draw [->] (j) -- (i);
\end{tikzpicture}
\item  $\omega_{ij}^4$: $a_i \sim a_j$:
\begin{tikzpicture}
\pgfmathsetmacro{\shift}{0.2ex}
\node[draw,circle] (i) at  (0,0) {i};
\node[draw,circle] (j) at  (2,0) {j};

\end{tikzpicture}
\end{itemize} 
In the preference graph, a Condorcet Paradox is represented by Strongly Connected Component (or circle). A strongly connected component of size 2 is considered as an indifference relationship, therefore the Condorcet Paradox is represented by circles of minimum size 3. A simple example of Condorcet circle is illustrated bellow:\\
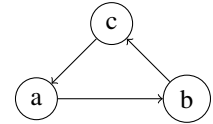
\begin{figure}[h]
\begin{center}
\begin{tikzpicture}
\node[draw,circle] (a) at  (0,0) {a};
\node[draw,circle] (b) at  (2,0) {b};
\node[draw,circle] (c) at  (1,1) {c};
\draw [->] (a) -- (b);
\draw [->] (b) -- (c);
\draw [->] (c) -- (a);
\end{tikzpicture}
\caption{Graphic representation of Condorcet Paradox}
\end{center}
\end{figure}

The Condorcet paradox does not respect the property of transitivity, which must be satisfied in the quasi order (definition~\ref{def:partialorder}). To avoid the Condorcet paradox, we have to cut an edge in the circle of size larger than 3. The fact of removing an edge between $a$ and $b$ is equivalent to replace the original relation between $a$ and $b$ by ``incomparability". In order to introduce as little knowledge as possible, we decide to remove the edge which is the closest to the relation ``incomparability". In our work, we choose Jousselme distance \cite{Jousselme200191} for distance measurement. Jousselme distance is considered as an reliable similarity measure between different mass functions~\cite{essaid:hal-01110349}. It considers coefficients on the elements composed by singletons. In this paper, Jousselme distance is denoted by $d_J$.
Hence distance between an alternative pair $(a_i,a_j)$ and ``incomparability" is denoted by $d_J(m_{ij},m^{\omega_4})$, 
where the mass function of incomparability $m^{\omega_4}$ is valued as $m^{\omega_4}(\omega_4)=1$.

Obviously, preferences over multiple alternatives without Condorcet paradox can be represented by a Directed Acyclic Graph (DAG) (note that cycles of 2 elements are tolerated in such graph because of the ``indifference" relation). To simplify the explanation, DAG with tolerance of 2-element-circle is denoted by $DAG_2$.

With Strongly Connected Components (SCC) search methods such as Tarjan's algorithm~\cite{Tarjan72depthfirst}, we firstly propose a ``naive algorithm" by iterating ``DAG detect-edge remove" process in algorithm~\ref{alg:naive}. Each iteration detects all the SCC of size larger than 2 as subgraphs (function SCC in algorithm~\ref{alg:naive}) and remove one edge closest to incomparability in each sub-graph (\textit{Loop Process} in algorithm~\ref{alg:naive}).
\begin{algorithm}[htb]
\caption{Naive $DAG_2$ Building Algorithm}
\label{alg:naive}
\begin{algorithmic}[1]
\renewcommand{\algorithmicrequire}{\textbf{Input:}}
 \renewcommand{\algorithmicensure}{\textbf{Output:}}
  \REQUIRE  a preference graph $G$ constructed on equation~\eqref{eq:decision} with edges valued by mass functions $edge.mass$
 \ENSURE  A $DAG$ graph
 \\ \textit{// Loop Process}
 \WHILE {$subgraphList$=\textit{SCC}($G$) is not empty}
 \FOR {$subgraph$ in $subgraphList$}
 	\STATE remove $edge$ in $subgraph.edges$ whose $d_J(edge.mass, m^{\omega_4})$ is the minimum
 \ENDFOR
 \ENDWHILE
 \\ \textit{// SCC search function}
 \\ \textbf{function} SCC($G$)
 \\ \hspace{\algorithmicindent} \textbf{Input:} directed graph $G$
 \\ \hspace{\algorithmicindent} \textbf{Output:} The sub-graphs of strongly connected components of size larger than 2 in $G$.

\end{algorithmic}
\end{algorithm}

\begin{figure}[h]
\centering
\begin{tikzpicture}
\node[draw, circle] (1) at  (0,0) {1};
\node[draw, circle] (2) at  (2,0) {2};
\node[draw, circle] (3) at  (1,0.5) {3};
\node[draw, circle] (4) at  (1,1.2) {4};
\node[draw, circle] (N) at  (1,2.3) {N};
\node at ($(4)!.6!(N)$) {\vdots};
\path[->] (2) edge (1);
\path[->] (1) edge (3);
\path[->] (3) edge (2);
\path[->] (1) edge (4);
\path[->] (4) edge (2);
\path[->] (1) edge (N);
\path[->] (N) edge (2);
\end{tikzpicture}
\caption{Nested circles}
\label{fig:nestedstructure}
\end{figure}
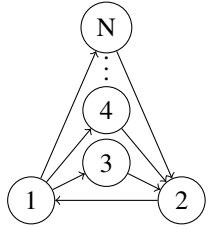

Since the SCC search function returns the largest strongly connected component found, this method loses its efficiency confronting SCC with nested circles. A simple structure of nested circles is illustrated in Figure~\ref{fig:nestedstructure}. 

Facing such structure of preferences, we propose here a more efficient algorithm to build a $DAG_2$ in an incremental way, described in algorithm~\ref{alg:incremental}. In this algorithm, all edges are ordered by their Jousselme distance to incomparability at the initialization phase (line 2). ``Incremental" means that the graph is built by adding edges one by one in an descending order of their distance to incomparability. Given an alternative pair $a_i,a_j$ with their predefined comparable (preference, inverse preference or indifference) relation $r$ calculated from the mass function, we check if the graph is still a $DAG_2$ with new edge between $a_i,a_j$ added. The checking process is based on a Depth-First-Search (DFS). Given two nodes $a_i$ and $a_j$ in graph $G$, function $DFS(G,a_i,a_j)$ returns the path length from $a_i$ to $a_j$.

More precisely, if a relation $r$ is a strict preference $a_i \succ a_j$, the DFS algorithm searching node $a_i$ starts from node $a_j$. If $a_i$ is found, a circle will appear if the edge $i\longrightarrow j$ is added. In such case, we replace the relation between $a_i,a_j$ by an incomparability relation (remove the edge) (line 6 and 7). Similarly if $r$ is inverse preference $a_i \prec a_j$, The DFS algorithm searches node $a_j$ from $a_i$ (line 8 and 9). 
However, the relation ``indifference" may hinder the temporal performance of this algorithm. If $r$ represent the indifference relation $a_i \approx a_j$, we have to apply DFS twice from $a_i$ to $a_j$ and from $a_j$ to $a_i$ (line 10 and 11). 
\begin{algorithm}[htb]
\caption{Incremental $DAG_2$ Building Algorithm}
\label{alg:incremental}
\begin{algorithmic}[1]
\renewcommand{\algorithmicrequire}{\textbf{Input:}}
 \renewcommand{\algorithmicensure}{\textbf{Output:}}
  \REQUIRE  All pairs $PAIRS$ and their mass functions $M$
  \ENSURE  A $DAG_2$ graph
 \\ \textit{// Initialization}:
 \STATE Initialize an empty graph $G$ 
 \STATE Ascending order all pairs $PAIRS$ by $d_J(m_{ij},m^{\omega_4})$ \cite{Jousselme200191}, stock in a stack $Stack$.
 \\ \textit{// Loop Process}
 \WHILE {$Stack$ is not empty}
 	\STATE pair $(a_i,a_j)$ = $Stack.pop()$
 	\STATE Add nodes $a_i$ and $a_j$ into graph $G$
	\IF{$a_i\succ a_j$}
		\STATE pathLength=$DFS(G,a_j,a_i)$
	\ELSIF{$a_i \prec a_j$ }
		\STATE pathLength=$DFS(G,a_i,a_j)$
	\ELSIF{$a_i\approx a_j$}
		\STATE pathLength=max($DFS(G,a_j,a_i)$, $DFS(G,a_i,a_j)$)
	\ENDIF
 	\IF{pathLength$\geqslant 2$}
 		\STATE Consider relation between $(a_i,a_j)$ as incomparability
 	\ELSE
 		\STATE add edge between $(a_i,a_j)$ calculated by $m_{ij}$ (equation~\eqref{eq:decision})
 	\ENDIF
 \ENDWHILE
 \\ \textit{// Recursive Function}
 \\ \textbf{function} DFS($G, v, n$):
	\STATE label v as discovered 
 	\IF {all successors of $v$ in $G$ are labeled as discovered}
 	\STATE return 0
 	\ELSE
 		\FOR {$w$ in non-discovered successors of $v$}
 			\IF{$w$ is $n$} 
 				\STATE return 1
 			\ELSE
	 			\STATE $len$=DFS($G,w,n$)
 				\IF{$len$ == 0}
 					\STATE return 0
 				\ELSE
 					\STATE return $len$ + 1
 				\ENDIF
 			\ENDIF
 		\ENDFOR
 	\ENDIF
\end{algorithmic}
\end{algorithm}

In a structure of nested circles containing $E$ edges, $V$ vertex and $N$ circles, the naive algorithm based on SCC search (algorithm~\ref{alg:naive}) can reach a temporal complexity of $\mathcal{O}(N(\lvert E\rvert+\lvert V\rvert))$ while the incremental algorithm (algorithm~\ref{alg:incremental}) has a temporal complexity of $\mathcal{O}(N)$.  

Although the incremental algorithm is efficient on nested circle its temporal performance degenerates when the preference structure has few nested circles or many indifference relations. The applicability of the two algorithms is demonstrated and discussed in the following section.

\section{Experiments}\label{sec:experiments}
In this section, we compare two preference fusion strategies and two algorithms for Condorcet paradox avoidance. The fusion strategies are evaluated from a numeric point of view while the algorithms are evaluated in terms fo temporal performance.
Lacking social network data from real world, the data used in our experiments were generated manually or randomly. 
The methods are implemented in python 3.5 on Fedora 24. The calculation concerning belief function theory is an adaptation of the R package ``ibelief" and graph manipulations are implemented with the help of python package ``networkx 1.11". The experimental platform is equipped with a CPU of Intel i7-6600U @2.6GHz $\times$4 and a memory of 16GB, while the programs run only on one of the CPU cores. 
\subsection{Preference fusion strategies}
\begin{figure}[h]
\centering
\begin{tikzpicture}
\node[draw, circle] (1) at  (0,0.5) {1};
\node[draw, circle] (2) at (1.5,0){2};
\node[draw, circle] (3) at (2.5,1.2){3};
\node[draw, circle] (4) at  (3.5,0) {4};
\node[draw, circle] (5) at (5,0.7){5};
\node[] (agent1) at(2.5,-0.5) {Agent 1};
\path[<->](1) edge (2);
\path[->](2) edge (3);
\path[->](1) edge (2);
\path[->](3) edge (4);
\path[->](3) edge (5);
\path[->](4) edge (5);
\end{tikzpicture}
\centering
\begin{tikzpicture}
\node[draw, circle] (1) at  (0,0.5) {1};
\node[draw, circle] (2) at (1.5,0){2};
\node[draw, circle] (3) at (2.5,1.2){3};
\node[draw, circle] (4) at  (3.5,0) {4};
\node[draw, circle] (5) at (5,0.7){5};
\node[] (agent2) at(2.5,-0.5) {Agent 2};
\path[<->](1) edge (2);
\path[->](4) edge (2);
\path[->](4) edge (5);
\path[->](3) edge (4);
\end{tikzpicture}
\centering
\begin{tikzpicture}
\node[draw, circle] (1) at  (0,0.5) {1};
\node[draw, circle] (2) at (1.5,0){2};
\node[draw, circle] (3) at (2.5,1.2){3};
\node[draw, circle] (4) at  (3.5,0) {4};
\node[draw, circle] (5) at (5,0.7){5};
\node[] (agent3) at(2.5,-0.5) {Agent 3};
\path[->](2) edge (1);
\path[->](2) edge (3);
\path[->](4) edge (2);
\path[->](5) edge (4);
\end{tikzpicture}
\caption{Preference order of three agents}
\label{fig:user3}
\end{figure}
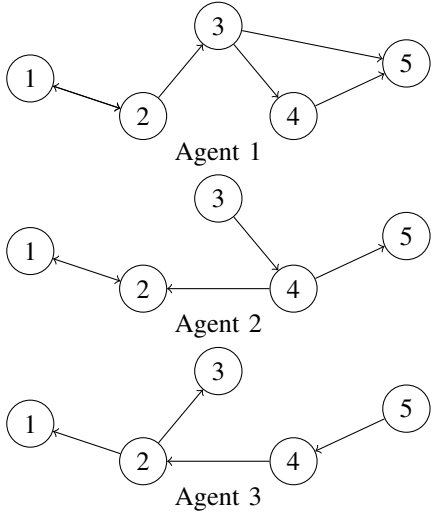

In this experiment, we firstly define preference structures of 3 agents as illustrated in figure~\ref{fig:user3}.
To simplify our experiments, the alternative pairs are always in an ascending order ({\em i.e.} $\forall a_i, a_j \in A$  $\Rightarrow i<j$).
The belief degrees for alternative pairs (2,3), (2,4) and (3,4) are specially given in table \ref{tab:degreevalue}. For the other alternatives, their belief degrees are set by default values given in table \ref{tab:defaultdegree}.
\begin{table}[H]
\centering
\caption{Belief degree values for alternative pairs (2,3), (2,4) and (3,4) }
\label{tab:degreevalue}
\begin{tabular}{|c|c|c|c|c|}
\hline 
user/pair & $\omega_1$ & $\omega_2$ & $\omega_3$ & $\omega_4$ \\ 
\hline 
User 1/(2,3) & 0.8 & 0.7 & 0.6 & 0.5 \\ 
\hline 
User 1/(2,4) & 0.4 & 0.1 & 0.3 & 0.6 \\ 
\hline 
User 1/(3,4) & 0.9 & 0.8 & 0.7 & 0.6 \\ 
\hline 
User 2/(2,3) & 0.5 & 0.4 & 0.6 & 0.9 \\ 
\hline 
User 2/(2,4) & 0.2 & 0.4 & 0.3 & 0.1 \\ 
\hline 
User 2/(3,4) & 0.9 & 0.8 & 0.1 & 0.7 \\ 
\hline 
User 3/(2,3) & 0.6 & 0.2 & 0.4 & 0.1 \\ 
\hline 
User 3/(2,4) & 0.3 & 0.5 & 0.2 & 0.1 \\ 
\hline 
User 3/(3,4) & 0.8 & 0.1 & 0.6 & 0.9 \\ 
\hline 
\end{tabular} 
\end{table}
\begin{table}[h]
\centering
\caption{Default belief degree values on 4 relations}
\label{tab:defaultdegree}
\begin{tabular}{|c|c|c|c|c|}
\hline 
relation & $\omega_1$ & $\omega_2$ & $\omega_3$ & $\omega_4$ \\ 
\hline 
preference & 0.8 & 0.2 & 0.3 & 0.1 \\ 
\hline 
inverse preference & 0.1 & 0.9 & 0.2 & 0.1 \\ 
\hline 
indifference & 0.3 & 0.3 & 0.7 & 0 \\ 
\hline 
incomparability & 0.1 & 0.1 & 0 & 0.9 \\ 
\hline 
\end{tabular} 
\end{table}
With the two fusion strategies A and B in Section \ref{subsec:fusionstrategies}, we get two different results illustrated in figure~\ref{fig:resultA} and \ref{fig:resultB}.
\begin{figure}[htb]
\centering
\begin{tikzpicture}
\node[draw, circle] (1) at  (0,0.5) {1};
\node[draw, circle] (2) at (1.5,0){2};
\node[draw, circle] (3) at (2.5,1.2){3};
\node[draw, circle] (4) at  (3.5,0) {4};
\node[draw, circle] (5) at (5,0.7){5};
\path[<->, thick,dash dot] (1) edge (2);
\path[->] (2) edge (3);
\path[->] (3) edge (4);
\path[->] (4) edge (2);
\path[->] (4) edge (5);
\end{tikzpicture}
\caption{Fusion result of strategy A}
\label{fig:resultA}
\end{figure}
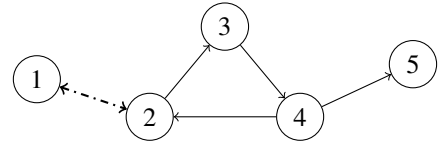

\begin{figure}[htb]
\centering
	
\caption{Fusion result of strategy B}
\label{fig:resultB}
\end{figure}
A different edge result between node 1 and 2 is high lighted by dashed dotted line. Focal elements in mass functions on edges are shown in table~\ref{tab:bbaA} and \ref{tab:bbaB}. Since in original data, mass function are zero on all non-singleton elements, the final mass function are still zero on union elements except ignorance (and empty set for strategy A).
\begin{table}[H]
\caption{Mass function values on edges of strategy A}
\label{tab:bbaA}
\centering
\begin{tabular}{|c|c|c|c|c|c|c|}
\hline 
pair & $\emptyset$ & $\omega_1$ & $\omega_2$ & $\omega_3$ & $\omega_4$ & $\Omega$ \\ 
\hline 
(1,2) & 0.17989 & 0.08620 & 0.19870 & 0.21626 & 0.01139 & 0.30755 \\ 
\hline 
(2,3) & 0.07858 & 0.17209 & 0.03612 & 0.08545 & 0.15812 & 0.46962 \\ 
\hline 
(2,4) & 0.08745 & 0.07225 & 0.12931 & 0.11762 & 0.12373 & 0.46962 \\ 
\hline 
(3,4) & 0.07858 & 0.17209 & 0.03612 & 0.08545 & 0.15812 & 0.46962 \\ 
\hline 
(4,5) & 0.20880 & 0.22056 & 0.15581 & 0.09589 & 0.03375  & 0.28519 \\ 
\hline 
\end{tabular} 
\end{table}
\begin{table}[H]
\caption{Mass function values on edges of strategy B}
\label{tab:bbaB}
\centering
\begin{tabular}{|c|c|c|c|c|c|c|}
\hline 
pair & $\emptyset$ & $\omega_1$ & $\omega_2$ & $\omega_3$ & $\omega_4$ & $\Omega$ \\ 
\hline 
(1,2) & 0 & 0.13975 & 0.23775 & 0.232 & 0.025 & 0.3655 \\ 
\hline 
(2,3) & 0 & 0.1765 & 0.05 & 0.10825 & 0.17 & 0.49525 \\ 
\hline 
(2,4) & 0 & 0.1 & 0.152 & 0.138 & 0.14875 & 0.46125 \\ 
\hline 
(3,4) & 0 & 0.1765 & 0.05 & 0.10825 & 0.17 & 0.49525 \\ 
\hline 
(4,5) & 0 & 0.241 & 0.234 & 0.152 & 0.06775  & 0.30525 \\ 
\hline 
\end{tabular} 
\end{table}
From this result, we observe that: 
\begin{enumerate}
\item Both combination strategies do not always return same results.
\item A significant difference between the two results exists in the value of empty set $\emptyset$.
\item A Condorcet Paradox appears in the fusion result.
\end{enumerate}

Concerning the empty set value $m^{\Omega_{ij}}(\emptyset)$, mass functions with plural focal elements imply auto-conflicts and cause an import value on $m^{\Omega_{ij}}(\emptyset)$ after conjunctive combination~\cite{Martin2008}. Moreover, with an increasing number of sources, $m^{\Omega_{ij}}(\emptyset)$ converges to 1 and other focal elements are no longer convincing.



In the Condorcet Paradox made up by alternatives 2, 3 and 4, the distance between the final combined mass functions associated to the edges of the circle to incomparability is given in table~\ref{tab:jousselmeDist}.

\begin{figure}[htb]
\centering
\begin{tikzpicture}
\node[draw, circle] (1) at  (0,0.5) {1};
\node[draw, circle] (2) at (1.5,0){2};
\node[draw, circle] (3) at (2.5,1.2){3};
\node[draw, circle] (4) at  (3.5,0) {4};
\node[draw, circle] (5) at (5,0.7){5};
\path[->] (2) edge (1);
\path[->] (3) edge (4);
\path[->] (4) edge (2);
\path[->] (4) edge (5);
\end{tikzpicture}
\caption{Result without Condorcet Paradox in strategy B}
\label{fig:resultDAGB}
\end{figure}
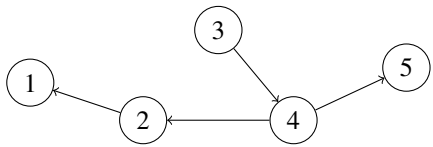

\begin{table}[H]\fontsize{11}{11}
\caption{Jousselme distance between alternative pair mass function to incomparability}
\label{tab:jousselmeDist}
\centering
\begin{tabular}{|c|c|c|}
\hline 
alternative pair & $d_J$ in Strategy A & $d_J$ in Strategy B \\
\hline 
(2,3) & 0.70798 & 0.60715 \\ 
\hline 
(3,4) & 0.77016 & 0.60739 \\ 
\hline 
(2,4) & 0.66928 & 0.64387 \\ 
\hline 
\end{tabular} 
\end{table} 
From table \ref{tab:jousselmeDist}, another difference between strategy A and B can be found. In strategy A, mass function of alternative pair $(2,4)$ is the closest to incomparability while in strategy B, the closest alternative pair to incomparability is $(2,3)$.
The final result is illustrated in figure \ref{fig:resultDAGA} and \ref{fig:resultDAGB}.
\begin{figure}[htb]
\centering
\begin{tikzpicture}
\node[draw, circle] (1) at  (0,0.5) {1};
\node[draw, circle] (2) at (1.5,0){2};
\node[draw, circle] (3) at (2.5,1.2){3};
\node[draw, circle] (4) at  (3.5,0) {4};
\node[draw, circle] (5) at (5,0.7){5};
\path[<->] (2) edge (1);
\path[->] (2) edge (3);
\path[->] (3) edge (4);
\path[->] (4) edge (5);
\end{tikzpicture}
\caption{Result without Condorcet Paradox in strategy A}
\label{fig:resultDAGA}
\end{figure}
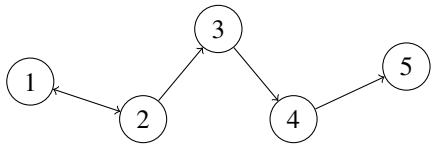

In Subsection~\ref{subsec:methodcondorcet}, we propose a method to overcome the Condorcet Paradox problem.

Considering the drawback caused by convergence of empty set value, we believe that strategy B outperforms strategy A, because strategy A does not scale with the number of agents. 

\subsection{Condorcet Paradox Avoidance}
\begin{figure}[h]
\centering
\includegraphics[width=8.5cm]{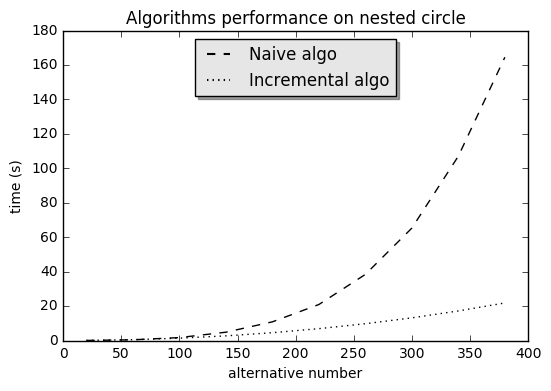}
\includegraphics[width=8.5cm]{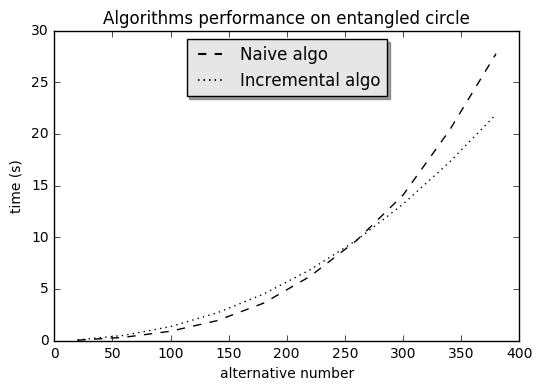}
\includegraphics[width=8.5cm]{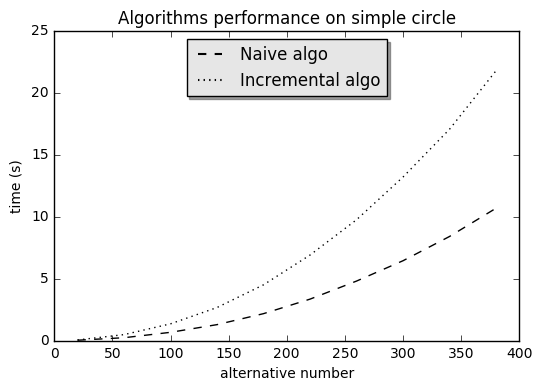}
\caption{Performances on different preference structures}
\label{fig:performanceonstructures}
\end{figure}

In this experiment, we evaluated the performance on three special preference structures: nested circles (figure~\ref{fig:nestedstructure}), entangled circles (figure~\ref{fig:entangled}) and non-nested structures with indifference relations (figure~\ref{fig:nonnested}). The mass functions associated to the preference relations are randomly generated adapting with entangled Condorcet paradox. The evaluation is based on the runtime with increasing number of nested circles $N$. As all mass functions are randomly generated following an uniform distribution from 0 to 1, we take the average value of 10 same tests to ensure the reliability of the result. In the experiments, alternative numbers range from 20 to 400 with an interval of 40.
\begin{figure}[H]
\begin{tikzpicture}[node distance=1.5cm]
\node[draw, circle] (1) {$1$};
\node[draw, circle] (2)[right of=1] {$2$};
\node[draw, circle] (3) [right of =2]{$3$};
\node[draw, circle] (4) [right of =3]{$4$};
\node[draw, circle] (5) [right of =4]{$5$};
\node[draw, circle,,scale=0.9] (N) [right of =5]{$N$};
\node at ($(5)!.5!(N)$) {\ldots};
\path[->](1) edge (2);
\path[->](2) edge (3);
\path[->](3) edge [bend right=90, looseness=1](1)
			 edge (4);
\path[->](4) edge [bend right=90, looseness=1](2)
			 edge (5);
\path[->](5) edge [bend left=90, looseness=1](3);
\path[->, loosely dotted](5) edge [bend left=90, looseness=1](N);
\path[->, loosely dotted](N) edge [bend left=90, looseness=1](4);
\end{tikzpicture}
\caption{Entangled Circles}
\label{fig:entangled}
\end{figure}
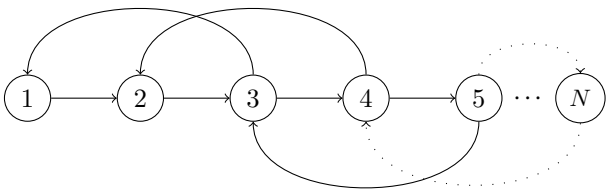
\begin{figure}[H]
\begin{tikzpicture}
\node[draw, circle] (1) at  (0,0) {1};
\node[draw, circle] (2) at (0.8,1.2){2};
\node[draw, circle] (3) at (1.6,0){3};
\node[draw, circle] (4) at  (3,0) {4};
\node[draw, circle] (5) at (3.8,1.2){5};
\node[draw, circle] (6) at (4.6,0){6};
\node at (5.5,0.5) {\ldots};
\node[draw, circle,scale=0.7] (N-2) at  (6.2,0) {N-2};
\node[draw, circle,scale=0.7] (N-1) at (7,1.2){N-1};
\node[draw, circle,scale=0.9] (N) at (7.8, 0){N};
\path[->](1) edge (2);
\path[->](2) edge (3);
\path[->](3) edge (1);
\path[<->](3) edge (4);
\path[->](4) edge (5);
\path[->](5) edge (6);
\path[->](6) edge (4);
\path[<->, dotted](6) edge (N-2);
\path[->](N-2) edge (N-1);
\path[->](N-1) edge (N);
\path[->](N) edge (N-2);
\end{tikzpicture}
\caption{Non-nested Circles}
\label{fig:nonnested}
\end{figure}
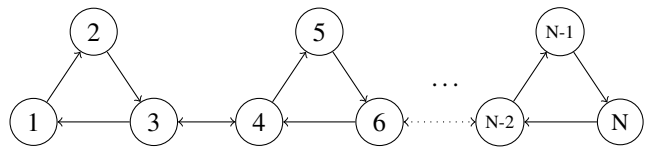
The performances related to the three preference structures are illustrated in figure~\ref{fig:performanceonstructures}.
We observe that the incremental algorithm outperforms the naive algorithm when the structures contain a great number of nested circles. However, for the structures with little nested circles but many indifference relations, the naive algorithm performs better. So the selection of Condorcet Paradox avoidance algorithm should be adapted to the structure of the preferences.

\section{Conclusion}\label{sec:conclusion}
In this paper, we introduced the problem of preference fusion with uncertainty degrees and proposed two belief-function-based strategies, one of which is applying the conjunctive rule on clusters, and which scales better with the number of sources. We also proposed a Condorcet Paradox avoidance method as well as an efficient DFS-based algorithm adapting to preference structure with nested circles. By comparing the temporal performance of the Condorcet Paradox avoidance algorithms on different types of preference structures, we noticed that the incremental algorithm is more efficient on nested structures while the naive algorithm is better on non-nested ones. Limited by our data sources, our experimental works were done on synthetic data. Furthermore, the algorithm for DAG construction can be applied in more general cases, other than those related to preference orders. In domains concerning directed graph with valued edges ({\em e.g.} telecommunication, social network analysis, {\em etc.}), algorithm \ref{alg:incremental} may find its usefulness. 

There are still more works left to explore. Since our experiments are executed on synthetic data, we were not able to propose evaluation methods measuring quality fusion results. In our work, we proposed methods to guarantee the ``transitivity" property, this property may help to reduce the number of ``incomparable" alternative pairs. Besides, in the Condorcet Paradox avoidance process, as the preference structure is altered so an information entropy measuring method is supposed to be proposed. Lastly, in real world, alternatives are often represented in multi-criteria forms. Conventional multi-critera aggregation methods ({\em e.g.} AHP method\cite{saaty2008decision}, PROMETHEE method\cite{brans1985note} and ELECTRE method\cite{roy1991outranking}) have been proved to be efficient. By combining these methods with belief-function theory, new aggregation methods for multi-criteria alternatives under uncertainty may be proposed as future work.


%

\appendices



\ifCLASSOPTIONcaptionsoff
  \newpage
\fi



\bibliographystyle{IEEEtran}
\bibliography{IEEEabrv,bibliography}
%



%





\end{document}